\documentclass[preprint,authoryear]{elsarticle}

\usepackage[utf8]{inputenc}
\usepackage[T1]{fontenc}
\usepackage{url}
\usepackage{booktabs}
\usepackage{amsfonts}
\usepackage{nicefrac}
\usepackage{microtype}
\usepackage[table]{xcolor}
\usepackage{apalike}
\usepackage{hyperref}

\usepackage{tikz}
\usetikzlibrary{shapes,arrows,positioning,fit,backgrounds}
\usepackage{graphicx}
\graphicspath{ {./} }

\usepackage[acronym]{glossaries}
\makeglossaries

\usepackage{float}
\usepackage{amsmath}
\usepackage{caption}
\usepackage{subcaption}
\usepackage{tcolorbox}

\newacronym{cot}{CoT}{Chain-of-Thought}
\newacronym{llm}{LLM}{Large Language Model}
\newacronym{vlm}{VLM}{Vision-Language Model}
\newacronym{lora}{LoRA}{Low-Rank Adaptation}
\newacronym{lvlm}{LVLM}{Large Vision-Language Model}
\newacronym{lcs}{LCS}{Longest Common Subsequence}
\newacronym{moe}{MoE}{Mixture-of-Experts}
\newacronym{sft}{SFT}{Supervised Fine-Tuning}
\newacronym{rl}{RL}{Reinforcement Learning}

\begin{document}

    \begin{frontmatter}

    \title{StoryMovie: A Dataset for Semantic Alignment of Visual Stories with Movie Scripts and Subtitles}

    \author[inesc,ist]{Daniel Oliveira}
    \ead{daniel.oliveira@inesc-id.pt}
    \author[inesc,ist]{David Martins de Matos}
    \ead{david.matos@inesc-id.pt}

    \address[inesc]{INESC-ID Lisboa, R. Alves Redol 9, 1000-029 Lisboa, Portugal}
    \address[ist]{Instituto Superior T\'{e}cnico, Universidade de Lisboa, Av. Rovisco Pais, 1049-001 Lisboa, Portugal}

    \begin{abstract}
        Visual storytelling models that correctly ground entities in images may still hallucinate semantic relationships,
        generating incorrect dialogue attribution, character interactions, or emotional states.
        We introduce StoryMovie, a dataset of 1,757 stories aligned with movie scripts and subtitles
        through \gls{lcs} matching.
        Our alignment pipeline synchronizes screenplay dialogue with subtitle timestamps,
        enabling dialogue attribution by linking character names from scripts to temporal positions from subtitles.
        Using this aligned content, we generate stories that maintain visual grounding tags
        while incorporating authentic character names, dialogue, and relationship dynamics.
        We fine-tune Qwen Storyteller3 on this dataset, building on prior work in visual grounding and entity re-identification.
        Evaluation using DeepSeek V3 as judge shows that Storyteller3 achieves an 89.9\% win rate
        against base Qwen2.5-VL 7B on subtitle alignment.
        Compared to Storyteller, trained without script grounding, Storyteller3 achieves
        48.5\% versus 38.0\%, confirming that semantic alignment progressively improves dialogue attribution
        beyond visual grounding alone.
    \end{abstract}


    \end{frontmatter}

    \section{Introduction}\label{sec:introduction}

    Visual storytelling, the generation of narratives from image sequences, represents a challenge at the intersection of computer vision
    and natural language processing with applications in entertainment, education, and journalism~\citep{Oliveira2024StoryGF}.
    Recent advances in \glspl{lvlm} have improved both image captioning~\citep{li2023blip, zhang2024groundhog}
    and visual narrative generation~\citep{tacl_a_00553, Yu_2021_CVPR}, yet current systems continue to produce
    stories that suffer from inconsistent character references, contradictory object relationships,
    and hallucinated content~\citep{Oliveira2024StoryGF}.

    Prior work has addressed visual grounding challenges in storytelling.
    GroundCap~\citep{Oliveira2025GroundCapAV} introduced ID-based grounding for single images using specialized tags
    that link textual references to detected visual entities.
    StoryReasoning~\citep{Oliveira2025StoryReasoningDU} extended this to multi-image contexts with cross-frame
    entity re-identification using visual similarity and face recognition, combined with \gls{cot}
    reasoning~\citep{wei2022chain, zhang2023multimodal} for structured narrative modeling.
    These approaches successfully reduce referential hallucinations where models fail to correctly identify
    or track visual entities across frames.
    However, visual grounding alone cannot prevent all types of hallucinations:
    even with perfect entity re-identification, models may correctly identify visual entities
    but incorrectly describe their relationships, emotions, or interactions relative to the actual narrative context.
    For example, a model might correctly ground two characters in an image but describe them as romantic partners
    when they are actually family members, or generate dialogue and attribute it to a character who did not speak in that scene.
    These semantic errors cannot be detected through visual analysis alone, as the visual appearance
    of characters provides no information about their actual relationships, the dialogue being spoken,
    or the narrative context established by the screenplay.

    Addressing this challenge requires grounding stories not only in visual content but also in authentic narrative sources.
    We introduce the StoryMovie dataset that aligns generated stories with actual
    movie scripts and subtitles, leveraging the high quality of professional screenwriting to provide
    ground-truth character relationships and dialogue attribution.
    The key insight is that movie scripts provide rich contextual information including character names, action lines, and emotional delivery cues,
    but lack precise temporal alignment with individual frames, while subtitles provide accurate timestamps
    for spoken dialogue but lack character attribution.
    By aligning these two sources, we create a temporally grounded representation of movie content
    that enables proper dialogue attribution.
    StoryMovie comprises 1,757 stories derived from movie sequences in StoryReasoning, each aligned with
    the corresponding screenplay segments and subtitle text.

    Building on this dataset, we develop Qwen Storyteller3, a model that inherits the structured \gls{cot}
    reasoning and visual grounding from prior training stages while adding semantic alignment with authentic
    movie narratives.
    This model represents the third stage of a progressive training approach:
    (1) basic visual grounding and \gls{cot} reasoning in Qwen Storyteller;
    (2) improved entity re-identification through contrastive reinforcement learning in Qwen Storyteller2; and
    (3) semantic alignment with movie scripts in Qwen Storyteller3.

    Beyond dialogue attribution, training on script-aligned data teaches the model a broader multimodal
    alignment: associating visual appearances with character names, mapping facial expressions and body language
    to screenplay delivery cues (e.g., ``\textit{(angrily)}'', ``\textit{(trembling)}''), and connecting
    scene compositions with action line descriptions.
    The screenplay thus serves as an external semantic anchor, allowing the model to draw on the screenwriter's
    intended characterization, emotional tone, and plot progression rather than hallucinating plausible
    but ungrounded narrative details from visual cues alone.

    The main contributions of this work are:
    (1) the StoryMovie dataset containing 1,757 stories with script-subtitle alignment that provides ground-truth
    semantic context for visual storytelling;
    (2) a script-subtitle alignment pipeline to synchronize screenplay content with subtitle timestamps for dialogue attribution; and
    (3) Qwen Storyteller3, a model with semantic alignment capabilities that generates
    stories with authentic dialogue and relationship dynamics.

    \section{Related Work}\label{sec:related-work}

    This section reviews relevant advances in visual storytelling, grounded visual narratives,
    script-subtitle alignment, and character-aware dialogue attribution.

    \subsection{Visual Storytelling}\label{subsec:visual-storytelling}

    Visual storytelling extends beyond image captioning by generating narratives that connect multiple images
    through causal and temporal relationships.
    Early approaches such as \citet{huang2016visual} used sequential recurrent architectures to generate stories
    from image sequences but struggled with maintaining character consistency and producing coherent narratives.
    TAPM~\citep{Yu_2021_CVPR} introduced transitional adaptation to align visual and textual information,
    while CharGrid~\citep{tacl_a_00553} implicitly modeled characters and their relationships across frames.
    HEGR~\citep{pmlr-v139-zheng21b} used hypergraphs to model relationships between scene elements,
    and PR-VIST~\citep{hsu2024prvist} introduced a plot-and-rework framework with iterative refinement.
    While these methods improve narrative quality, they do not explicitly track object identities
    across frames or ground story elements in authentic narrative sources.
    Because these approaches generate narratives from visual content alone, they are susceptible
    to factually incorrect character interactions---a model may infer romantic tension
    between two characters who are actually family members, or fabricate dialogue that contradicts the actual
    screenplay.
    Our approach addresses this limitation by grounding generated stories in authentic screenplay content,
    providing the model with ground-truth character relationships and dialogue that resolve ambiguities
    inherent in visual-only interpretation.

    \subsection{Grounded Visual Storytelling}\label{subsec:grounded-visual-storytelling}

    GroundCap~\citep{Oliveira2025GroundCapAV} introduced ID-based visual grounding for image captioning
    using specialized tags (\texttt{<gdo>}, \texttt{<gda>}, \texttt{<gdl>}) that link textual references
    to detected objects, actions, and locations within individual images.
    The dataset contains 52,016 images from 77 movies with both automated and human-refined captions.
    StoryReasoning~\citep{Oliveira2025StoryReasoningDU} extended this to multi-image contexts with 4,178 stories
    that maintain cross-frame entity consistency through re-identification using SigLIP visual
    embeddings~\citep{zhai2023sigmoid} and ArcFace face recognition~\citep{arcface2018}.
    The dataset includes structured \gls{cot} analyses with tabular representations for characters, objects,
    settings, and narrative progression.
    Qwen Storyteller, fine-tuned on StoryReasoning, demonstrated a 12.3\% reduction in hallucinations
    and 31.0\% improvement in creativity compared to the base model.
    Despite these improvements, models trained on StoryReasoning generate stories without connection
    to ground-truth narrative content, as the training stories were produced by \glspl{vlm}
    describing visual content without access to actual movie scripts.
    A direct consequence is that character names are vague or generic, drawn from a small pool
    of common names (e.g., ``John'', ``Sarah'', ``David'') that the \gls{vlm} defaults to
    when it has no way to determine the actual identities of the people it sees.
    Our StoryMovie approach resolves this by providing actual character names and dialogue from screenplays,
    directly addressing the naming and dialogue attribution limitations of visual-only story generation.

    \subsection{Script-Subtitle Alignment}\label{subsec:script-subtitle-alignment-related}

    Script-subtitle alignment addresses the challenge of synchronizing screenplay content with video.
    Scripts provide rich semantic information including character names, dialogue, and scene descriptions,
    but lack temporal grounding, while subtitles provide accurate timestamps but lack character attribution.
    \citet{everingham2006hello} pioneered the use of Dynamic Time Warping to align movie scripts with subtitles
    for automatic character naming in TV videos, demonstrating that alignment enables determining
    who says what and when.
    \citet{cour2008movie} extended this for parsing movies into hierarchical shot and scene structures
    using a unified generative model.
    The \gls{lcs} algorithm~\citep{bergroth2000survey} finds the longest subsequence common to two sequences.
    \citet{stanislav2012unsupervised} applied \gls{lcs} for unsupervised subtitle synchronization,
    using keyword spotting for approximate alignment followed by \gls{lcs} to determine
    the best alignment between audio transcripts and subtitle text.
    Our approach similarly applies \gls{lcs}-based token matching to synchronize
    screenplay dialogue with subtitle timestamps, enabling dialogue attribution.

    \subsection{Character-Aware Dialogue Attribution}\label{subsec:character-dialogue-attribution}

    Identifying which characters speak which lines is fundamental to understanding movie narratives.
    \citet{huh2024characteraware} advanced character-aware subtitling by integrating
    speech recognition, speaker diarization, and character recognition using audio-visual cues,
    employing \gls{llm} reasoning on text transcription to determine speakers
    from temporal context within scenes.
    CHATTER~\citep{chatter2025} introduced a character attribution dataset for narrative understanding
    using movies as the source, creating benchmarks for character identification and quote attribution
    by leveraging cast lists and character information.

    \glsreset{cot}
    \section{Background}\label{sec:background}

    This work builds on a progressive pipeline for visual storytelling.
    This section summarizes the prior stages that provide the foundation for the StoryMovie dataset
    and Qwen Storyteller3.

    \subsection{GroundCap: Single-Image Grounding}\label{subsec:groundcap-background}

    GroundCap~\citep{Oliveira2025GroundCapAV} established ID-based grounding for single-image captioning.
    The dataset contains 52,016 images from 77 MovieNet movies, each paired with captions that use
    specialized XML tags to link textual references to visual entities: \texttt{<gdo>} for objects,
    \texttt{<gda>} for actions, and \texttt{<gdl>} for locations.
    Each tag contains a unique identifier linking it to a specific bounding box in the image.
    This approach enables precise evaluation of visual-textual correspondence through the gMETEOR metric,
    which combines language quality assessment with grounding accuracy.

    \subsection{StoryReasoning: Multi-Image Grounding}\label{subsec:storyreasoning-background}

    StoryReasoning~\citep{Oliveira2025StoryReasoningDU} extended single-image grounding to sequential contexts.
    The dataset contains 4,178 stories derived from GroundCap images organized into temporally coherent sequences
    of at least 5 frames from the same movie.
    Cross-frame entity re-identification uses SigLIP~\citep{zhai2023sigmoid} visual embeddings
    for general objects and ArcFace~\citep{arcface2018} face recognition for characters,
    with additional \texttt{<gdi>} tags enabling temporal alignment across frames.

    Each story includes a structured \gls{cot} analysis with tabular representations covering
    character identification, object detection, setting descriptions, and narrative progression.
    Qwen Storyteller, fine-tuned on this dataset, performs end-to-end object detection, re-identification,
    and landmark detection while maintaining consistent references throughout generated stories.
    Evaluation showed a reduction from 4.06 to 3.56 hallucinations per story (-12.3\%) and
    improved creativity from 2.58 to 3.38 (+31.0\%) compared to the base Qwen2.5-VL 7B model,
    with reference alignment win rates above 72\% across all reference types
    (Table~\ref{tab:storyteller-vs-base}).

    Despite these improvements in visual grounding and entity tracking, models trained on StoryReasoning
    generate stories without access to ground-truth narrative content.
    The training stories describe what happens visually but have no connection to an actually plausible narrative,
    meaning models may correctly identify visual entities but incorrectly describe their relationships,
    dialogue, or emotional states.

    \section{The StoryMovie Dataset}\label{sec:storymovie-dataset}

    We extend StoryReasoning with StoryMovie, incorporating movie scripts and subtitles
    to provide ground-truth semantic context for story generation.
    The dataset consists of 1,757 stories derived from StoryReasoning by selecting the subset
    of stories whose source movies have both available scripts and subtitles.
    Each story maintains the visual grounding tags from StoryReasoning while being additionally
    aligned with the corresponding screenplay and subtitles.
    The dataset is randomly split into 1,494 training samples and 263 test samples (85\%/15\%),
    used for fine-tuning (\S\ref{sec:qwen-storyteller3}) and evaluation (\S\ref{sec:evaluation-results}), respectively.

\begin{figure*}[!htbp]
        \centering
        \includegraphics[width=0.8\textwidth]{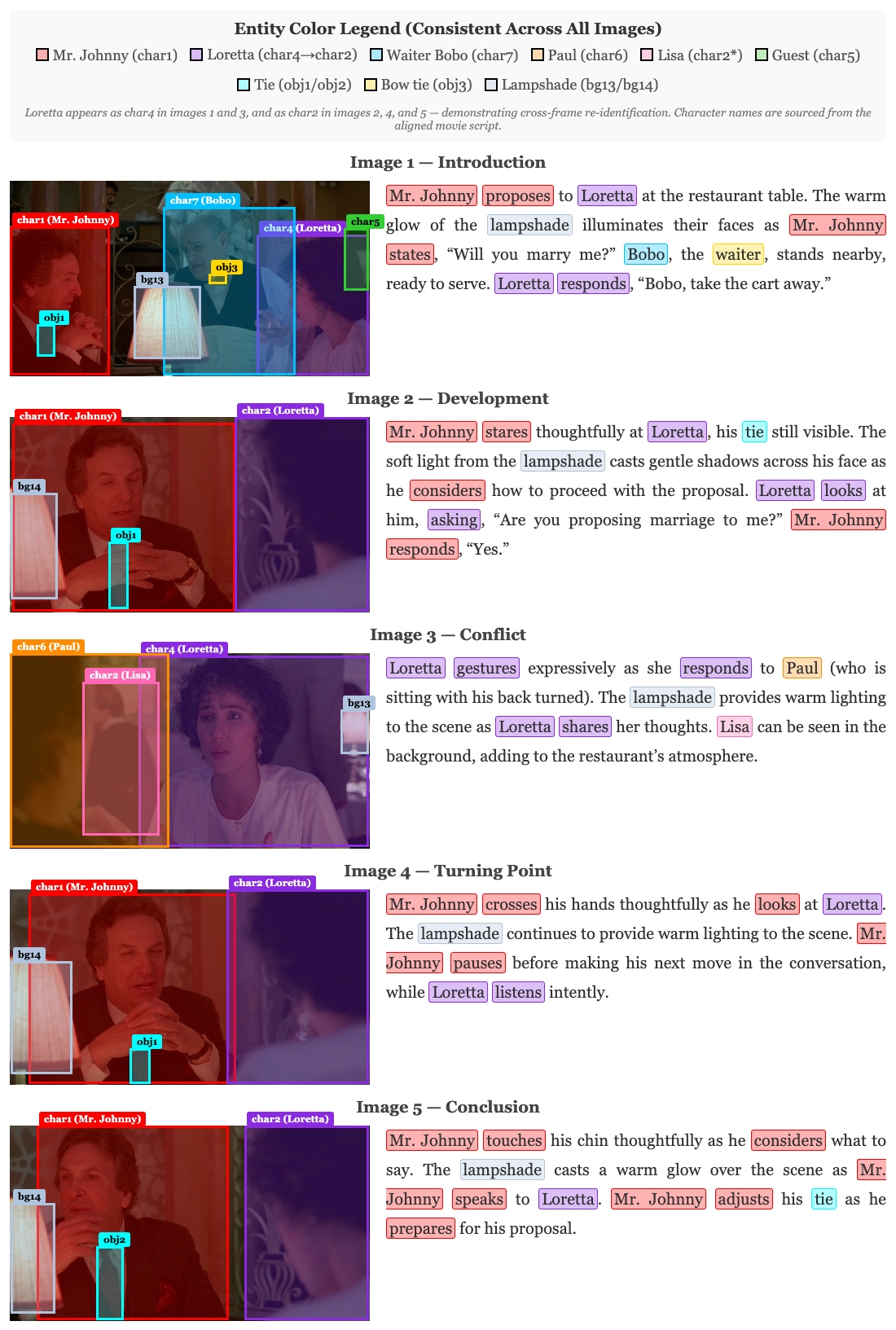}
        \caption{Example story from the StoryMovie dataset showing entity grounding with script-aligned dialogue.
        Character names from the screenplay (e.g., Mr.\ Johnny, Loretta) are linked to visual entities,
        and dialogue is attributed based on script-subtitle alignment.
        Loretta is re-identified across frames, appearing as char4 in images 1 and 3
        and as char2 in images 2, 4, and 5.}
        \label{fig:example-sample}
    \end{figure*}

    The stories (in English) average 674.81 words ($\sigma = 445.24$) and contain an average of 119.54 entity references,
    distributed across character mentions (32.64), object references (4.70), setting descriptions (32.65),
    and action references (49.55).
    Each story contains an average of 9.21 distinct characters.
    A complete example from our dataset is shown in Figure~\ref{fig:example-sample}, which illustrates how
    entities are grounded across a five-image sequence with script-aligned dialogue attribution.

    \subsection{Script-Subtitle Alignment}\label{subsec:script-subtitle-alignment}

    Scripts provide character names, dialogue, action lines, and emotional delivery cues but lack temporal alignment.
    Subtitles provide timestamps but lack character attribution.
    Aligning these sources creates a temporally grounded representation of movie content.

    Our alignment uses token-based matching to synchronize screenplay dialogue with subtitle text.
    We parse the screenplay to extract structured elements: scene headings (e.g., ``INT./EXT.''),
    character names (centered, uppercase), dialogue (indented text under character), actions (minimal indentation),
    and emotional delivery cues (e.g., ``\textit{(angrily)}'' or ``\textit{(softly)}'') that ground character emotional states in the intended tone of the screenplay.
    Action lines describe physical behavior, atmosphere, and non-verbal interactions beyond what is visually apparent in individual frames.
    Both screenplay dialogue and subtitle text are tokenized for matching.

    The alignment algorithm applies \gls{lcs} matching~\citep{bergroth2000survey} to identify corresponding
    dialogue blocks between screenplay and subtitles.
    Once a match is identified, the algorithm extends bidirectionally until the speaker changes
    or a scene break occurs, capturing complete dialogue sequences.
    Subtitle timestamps are assigned to aligned screenplay segments.
    For each story in StoryReasoning, we determine the temporal range covered by its frames
    using MovieNet annotations and extract the aligned screenplay segments that overlap with this range.

    \subsection{Story Generation with Script Grounding}\label{subsec:story-generation-script-grounding}

    We generate grounded stories using Qwen3-VL-235B-A22B~\citep{qwen3vl2025},
    a \gls{moe} \gls{vlm} that receives the original images, the structured \gls{cot} from StoryReasoning,
    the aligned script segment, and subtitle text.
    The \gls{vlm} maps visual entities to screenplay characters:
    if the \gls{cot} identifies ``char1'' as ``a woman in a blue dress''
    and the screenplay indicates ``Mary'' is speaking, the model associates
    ``char1'' with ``Mary'' based on visual context and dialogue cues.

    Beyond dialogue, the model receives the full screenplay context---action lines and delivery cues---which enables 
    the generation of stories that describe character emotional states grounded in the intended tone of the screenplay 
    rather than inferred solely from visual appearance, and that capture atmosphere and non-verbal interactions 
    described in the action lines.

    Generated stories maintain the grounding tag structure from StoryReasoning, using XML tags to link
    narrative elements to visual entities.
    The content reflects the actual movie narrative with correct character names,
    dialogue attribution, and proper relationship descriptions from the screenplay.
    The StoryMovie dataset is publicly available at \url{https://huggingface.co/datasets/daniel3303/StoryMovieScript}
    under the same license as MovieNet.

    \section{Qwen Storyteller3}\label{sec:qwen-storyteller3}

    Building on the improved entity re-identification capabilities achieved through contrastive \gls{rl},
    we further fine-tune the model on the StoryMovie dataset to achieve semantic alignment with movie scripts
    and subtitles.
    This additional training phase teaches the model to generate stories that respect the actual character
    relationships, dialogue, and plot developments as defined in the original screenplay.

    \subsection{Training Configuration}\label{subsec:storyteller3-training}

    We fine-tune the contrastive \gls{rl} model on the StoryMovie dataset using \gls{sft}.
    We experimented with \gls{lora}~\citep{hu2022lora} ranks 8, 32, 128, and 512, selecting rank 32
    (alpha 64) based on a heuristic combining the number of well-structured stories the model
    could generate on the test set with entity re-identification performance,
    ensuring the model retains skills acquired during previous training phases.
    Training uses a learning rate of $2 \times 10^{-4}$ with cosine annealing~\citep{loshchilov2017sgdr}
    and AdamW optimizer~\citep{loshchilov2018decoupled}, with an effective batch size of 32
    achieved through gradient accumulation.
    The model is trained for 3 epochs on the StoryMovie train split.

    \subsection{Model Capabilities}\label{subsec:storyteller3-capabilities}

    Qwen Storyteller3 inherits all capabilities from its predecessors while adding semantic alignment:
    from Qwen Storyteller, it maintains coherent multi-image stories with
    structured \gls{cot} reasoning and cross-frame entity grounding;
    from the contrastive \gls{rl} training, it preserves improved entity re-identification and pronoun grounding.
    The StoryMovie training adds the ability to generate stories that reflect authentic screenplay 
    content---character names, dialogue, emotional tone, and narrative relationships---rather than relying solely on visual inference.

    Qwen Storyteller3 is publicly available at \url{https://huggingface.co/daniel3303/QwenStoryteller3}
    for academic research.

    \section{Evaluation Results}\label{sec:evaluation-results}

    To evaluate the effectiveness of semantic alignment training, we compare Qwen Storyteller3 against
    both the base Qwen2.5-VL 7B model and the baseline using pairwise preference evaluation.
    We employ DeepSeek V3~\citep{deepseekai2025deepseekv3} as an \gls{llm} judge to perform pairwise comparisons
    on 341 test samples across three independent runs, evaluating against three reference types:
    subtitles (measuring dialogue attribution accuracy), descriptions (measuring visual scene understanding),
    and aligned screenplay content, referred to as synopsis (measuring broader narrative alignment).

    \subsection{Comparison with Base Model}\label{subsec:comparison-base-model}

    Table~\ref{tab:semantic-alignment-base} presents the pairwise comparison between Qwen Storyteller3
    and the base Qwen2.5-VL 7B model.

    \begin{table*}[htbp]
        \centering
        \caption{Semantic alignment evaluation: Ours (Qwen Storyteller3) vs.\ Qwen2.5-VL 7B (base model).
        Values represent win rates from pairwise preference evaluation.
        Results are mean $\pm$ standard deviation across three independent evaluation runs.}
        \label{tab:semantic-alignment-base}
        \begin{tabular}{lccc}
            \toprule
            Reference Type & Ours & Qwen2.5-VL 7B & Ties \\
            \midrule
            Subtitles      & \cellcolor{green!20}89.9\% $\pm$ 1.4 & 3.5\% $\pm$ 0.8  & 6.5\% $\pm$ 0.4  \\
            Description    & \cellcolor{green!20}63.4\% $\pm$ 0.9 & 4.1\% $\pm$ 0.9  & 32.5\% $\pm$ 1.2 \\
            Synopsis       & \cellcolor{green!20}87.6\% $\pm$ 1.1 & 6.8\% $\pm$ 0.7  & 5.7\% $\pm$ 0.5  \\
            \bottomrule
        \end{tabular}
    \end{table*}

    Qwen Storyteller3 demonstrates substantial improvements over the base model across all reference sources.
    On subtitle alignment, which directly measures dialogue attribution accuracy, Qwen Storyteller3 achieves
    an 89.9\% win rate compared to only 3.5\% for the base model, representing an 86.4 percentage point advantage.
    On description alignment, where the tie rate is highest (32.5\%), Qwen Storyteller3 wins 63.4\% of comparisons
    versus only 4.1\% for the base model, a fifteen-fold advantage.
    Synopsis alignment shows a similar pattern with an 87.6\% win rate versus 6.8\%.

    \subsection{Comparison with Qwen Storyteller}\label{subsec:comparison-qwen-storyteller}

    Table~\ref{tab:semantic-alignment-storyteller} presents the comparison between Qwen Storyteller3
    and Qwen Storyteller (trained without script alignment).

    \begin{table*}[htbp]
        \centering
        \caption{Semantic alignment evaluation: Ours (Qwen Storyteller3) vs.\ Baseline (Qwen Storyteller, trained without script alignment).
        Values represent win rates from pairwise preference evaluation.
        Results are mean $\pm$ standard deviation across three independent evaluation runs.}
        \label{tab:semantic-alignment-storyteller}
        \begin{tabular}{lccc}
            \toprule
            Reference Type & Ours & Baseline & Ties \\
            \midrule
            Subtitles      & \cellcolor{green!20}48.5\% $\pm$ 1.4 & 38.0\% $\pm$ 1.6 & 13.5\% $\pm$ 0.5 \\
            Description    & \cellcolor{green!20}35.5\% $\pm$ 1.7 & 15.2\% $\pm$ 1.3 & 49.2\% $\pm$ 1.0 \\
            Synopsis       & \cellcolor{green!20}42.7\% $\pm$ 1.5 & 28.5\% $\pm$ 1.4 & 28.8\% $\pm$ 0.8 \\
            \bottomrule
        \end{tabular}
    \end{table*}

    The results show consistent improvements from semantic alignment training.
    On subtitle alignment, Qwen Storyteller3 achieves a 48.5\% win rate compared to 38.0\% for
    Qwen Storyteller, representing a 10.5 percentage point advantage.
    This improvement demonstrates that the StoryMovie training helps the model generate dialogue
    that better matches the actual movie content.
    The high tie rate (49.2\%) on description alignment suggests that both models perform similarly on
    general scene description, which is expected because both models share the same visual grounding foundation
    from the StoryReasoning training pipeline.
    StoryReasoning stories emphasize visual descriptions with minimal dialogue,
    while StoryMovie stories focus more on dialogue content, explaining the stronger gains on dialogue-related references.
    The synopsis comparison shows a similar pattern, with Qwen Storyteller3 winning 42.7\% of comparisons
    versus 28.5\% for Qwen Storyteller, reflecting better alignment with the broader narrative context
    and character relationships as defined in movie scripts.

    For context, Table~\ref{tab:storyteller-vs-base} presents the reference alignment evaluation
    of the baseline against the base Qwen2.5-VL 7B model from prior work~\citep{Oliveira2025StoryReasoningDU},
    using the same pairwise preference methodology.
    The Baseline already achieves strong win rates (72--86\%) over the base model,
    confirming that the visual grounding training provides a solid foundation
    upon which semantic alignment further improves.

    \begin{table*}[htbp]
        \centering
        \caption{Reference alignment evaluation: Baseline (Qwen Storyteller) vs.\ Qwen2.5-VL 7B from prior work.
        Values represent win rates from pairwise preference evaluation.
        Results are mean $\pm$ standard deviation across three independent evaluation runs.}
        \label{tab:storyteller-vs-base}
        \begin{tabular}{lccc}
            \toprule
            Reference Type & Baseline & Qwen2.5-VL 7B & Ties \\
            \midrule
            Subtitles      & \cellcolor{green!20}86.4\% $\pm$ 0.9 & 12.2\% $\pm$ 1.2 & 1.4\% $\pm$ 0.3 \\
            Description    & \cellcolor{green!20}72.4\% $\pm$ 1.2 & 23.6\% $\pm$ 1.8 & 4.0\% $\pm$ 0.7 \\
            Synopsis       & \cellcolor{green!20}86.0\% $\pm$ 1.2 & 12.2\% $\pm$ 0.8 & 1.9\% $\pm$ 0.5 \\
            \bottomrule
        \end{tabular}
    \end{table*}

    These results confirm that script-aligned training is necessary to reduce higher-level hallucinations
    that visual grounding alone cannot address.
    The script-aligned training also provides emotional and atmospheric grounding through screenplay delivery cues and action lines;
    while our evaluation does not isolate emotional accuracy as a separate dimension, the synopsis reference
    implicitly captures some of this benefit.

    \subsection{Qualitative Analysis}\label{subsec:qualitative-analysis}

    We manually examined randomly sampled stories from both models on the test set.
    The baseline shows repetition, often recycling the same paragraph across images
    with minimal variation---in extreme cases repeating identical sentences with only the time reference
    changed (``days'', ``weeks'', ``months'').
    It relies heavily on abstract emotional narration (``the wall seemed to echo the tension in his chest'',
    ``the weight of recent events pressed heavily on his shoulders'') rather than describing
    concrete actions visible in the frames.
    Character names are drawn from a small pool of common English names (``John'', ``Sarah'', ``Mike''), 
    and dialogue is either absent or entirely fabricated and abstract (e.g., ``I can't believe this is happening'', ``We need to get out of here'').

    Our model produces more concise stories that are better anchored in visual content,
    describing specific actions (``lay in bed'', ``sat in chair'', ``stood near door'')
    that correspond to what is depicted in each frame.
    It occasionally includes dialogue and shows improved alignment with the narrative context
    established by the screenplay, which explains the stronger gains on subtitle and synopsis references.
    Our model also exhibits greater variety in character naming compared to the baseline.
    However, when script-subtitle alignment is weak for a scene---when subtitles are sparse
    or temporal overlap is minimal---the model falls back on visual-only inference.

    \subsection{Question-Answering Evaluation}\label{subsec:qa-evaluation}

    To complement the pairwise preference evaluation, we introduce a question-answering evaluation
    that directly measures factual accuracy.
    For each test story with script alignment, an \gls{llm} (GPT-5) generates three
    multiple-choice comprehension questions from the aligned portion of movie script, one per category:
    emotional state, action, and relationship.
    Questions use descriptive references (e.g., ``the older man'', ``the woman in red'')
    rather than character names, and each has three options with exactly one correct answer.
    The same \gls{llm} then answers each question using only the generated story text,
    and the selected answer is compared against the script-derived ground truth.
    This measures whether the story faithfully captures the factual content of the scene
    across three complementary dimensions.

    \begin{table}[htbp]
        \centering
        \caption{Factual accuracy evaluation via script-derived Q\&A.
        Accuracy (\%) on 38 test stories (114 questions total, 3 per story).}
        \label{tab:qa-accuracy}
        \begin{tabular}{lcccc}
            \toprule
            Model & Overall & Emot. & Action & Relat. \\
            \midrule
            Ours     & \cellcolor{green!20}93.9 & \cellcolor{green!20}89.5 & \cellcolor{green!20}97.4 & \cellcolor{green!20}94.7 \\
            Baseline & 63.2 & 65.8 & 68.4 & 55.3 \\
            \bottomrule
        \end{tabular}
    \end{table}

    Table~\ref{tab:qa-accuracy} shows that our model achieves 93.9\% overall accuracy
    compared to 63.2\% for the baseline, a 30.7 percentage point improvement.
    The largest gap appears on relationship questions (94.7\% vs.\ 55.3\%),
    consistent with the paper's central thesis that visual-only models struggle most
    with character relationships and interactions that are not directly observable from images.
    Action accuracy is near-perfect at 97.4\%, confirming that script alignment
    helps ground actions in authentic scene content.
    The baseline's weakest category is relationships, which aligns with the observation
    that inferring character dynamics from visual content alone is unreliable.

    \section{Limitations}\label{sec:limitations}

    Several limitations should be acknowledged.
    First, the research relies on movie-derived content, which introduces biases toward
    cinematic composition, professional lighting, and conventional narrative structures.
    Generalization to diverse visual sources such as personal photo albums, social media imagery,
    or surveillance footage remains to be validated.

    Second, the StoryMovie dataset contains 1,757 stories, a relatively modest size compared
    to large-scale vision-language datasets.
    While our results demonstrate consistent improvements, scaling the alignment pipeline to more movies
    and screenplays could yield further gains.

    Third, the evaluation relies on \gls{llm}-based judging with DeepSeek V3, which may introduce
    its own biases toward certain narrative styles or linguistic patterns.
    However, the consistency of results across multiple reference types and the alignment with
    expected patterns (stronger gains on dialogue-related metrics) provide confidence in the findings.

    Finally, the current work focuses exclusively on English-language content,
    limiting cross-cultural applicability.
    Future work should incorporate multilingual datasets and cross-cultural narrative analysis
    to broaden the scope of visual storytelling research.

    \section{Conclusions}\label{sec:conclusions}

    This work addressed higher-level semantic hallucinations in visual storytelling, where models correctly identify
    visual entities but incorrectly describe their relationships, dialogue, or emotional states.
    We introduced the StoryMovie dataset, containing 1,757 stories aligned with actual movie scripts and subtitles
    through a script-subtitle synchronization pipeline using \gls{lcs} matching.
    This dataset enables training on narratives grounded in authentic movie content, teaching models to respect
    actual character relationships, dialogue attribution, and plot developments as defined in the original screenplay.

    By fine-tuning the model on the StoryMovie dataset, we created Qwen Storyteller3, which demonstrates
    substantial improvements in semantic alignment.
    Evaluation shows that Qwen Storyteller3 achieves an 89.9\% win rate against the base Qwen2.5-VL 7B model
    on subtitle alignment, and a 48.5\% win rate against Qwen Storyteller (versus 38.0\% losses),
    confirming that semantic alignment training effectively reduces
    dialogue attribution errors that visual grounding alone cannot address.

    The progressive training approach across the Qwen Storyteller model family---from basic visual grounding,
    through contrastive \gls{rl} for entity re-identification, to semantic alignment with movie scripts---establishes
    a framework for addressing multiple types of hallucinations in visual storytelling.
    Each training stage builds upon the capabilities of previous stages while addressing distinct limitations,
    resulting in models that generate stories with accurate visual grounding, proper entity tracking,
    and semantically aligned dialogue and character relationships.

    Future work includes expanding to diverse visual sources beyond movies, developing unified training
    approaches that combine all data sources, and extending the framework to multilingual settings.

    \section*{Acknowledgments}

    Daniel Oliveira is supported by a scholarship granted by Funda\c{c}\~{a}o para a Ci\^{e}ncia e a Tecnologia (FCT),
    with reference 2021.06750.BD. Additionally, this work was supported by Portuguese national funds
    through FCT, with reference UIDB/50021/2020.

\end{document}